\documentclass[runningheads]{llncs}

\usepackage{eccv}

\usepackage{eccvabbrv}

\usepackage{graphicx}
\usepackage{booktabs}

\usepackage[accsupp]{axessibility}  

\usepackage[pagebackref,breaklinks,colorlinks]{hyperref}

\usepackage{orcidlink}

\begin{document}

\title{EffiVED:Efficient Video Editing via Text-instruction Diffusion Models} 


\author{Zhenghao Zhang\inst{1} \and
Zuozhuo Dai\inst{1}  \and Long Qin\inst{1} and 
Weizhi Wang\inst{1}}


\institute{Alibaba Group}

\maketitle

\begin{abstract}

Large-scale text-to-video models have shown remarkable abilities, but their direct application in video editing remains challenging due to limited available datasets. Current video editing methods commonly require per-video fine-tuning of diffusion models or specific inversion optimization to ensure high-fidelity edits.
In this paper, we introduce EffiVED, an efficient diffusion-based model that directly supports instruction-guided video editing. To achieve this, we present two efficient workflows to gather video editing pairs, utilizing augmentation and fundamental vision-language techniques. These workflows transform vast image editing datasets and open-world videos into a high-quality dataset for training EffiVED. Experimental results reveal that EffiVED not only generates high-quality editing videos but also executes rapidly. Finally, we demonstrate that our data collection method significantly improves editing performance and can potentially tackle the scarcity of video editing data. Code can be found at \href{https://github.com/alibaba/EffiVED}{https://github.com/alibaba/EffiVED}.

\end{abstract}

\section{Introduction}
\label{sec:intro}

The advent of text-to-image diffusion models\cite{DBLP:conf/cvpr/RombachBLEO22,DBLP:journals/corr/abs-2307-01952,DBLP:journals/corr/abs-2207-12598,DBLP:conf/icml/NicholDRSMMSC22} has propelled the advancement of text-driven video editing\cite{DBLP:conf/iccv/AbdalQW19, DBLP:conf/cvpr/AvrahamiLF22,DBLP:conf/eccv/CrowsonBKSHCR22,DBLP:journals/jmlr/HoSCFNS22}.  For example, Tune-A-Video\cite{wu2023tune} exemplifies this by initially fine-tuning a diffusion model with an input video and corresponding text description to establish the correspondence between the two modalities. Video-P2P\cite{DBLP:journals/corr/abs-2303-04761} optimizes a shared unconditional embedding to attain precise video inversion and utilizes attention swap in diffusion model for detailed editing.
On another front, CoDeF\cite{DBLP:journals/corr/abs-2308-07926} edits canonical image and propagates the changes temporally using deformation field extracted from a neural video field. This method delivers superior temporal coherence along with high-fidelity synthesized frames compared to the aforementioned methods. However, these methods all entail considerable computational costs because they require tailored processing for each video.

\begin{figure*}[t]
    \centering
    \includegraphics[width=\textwidth]{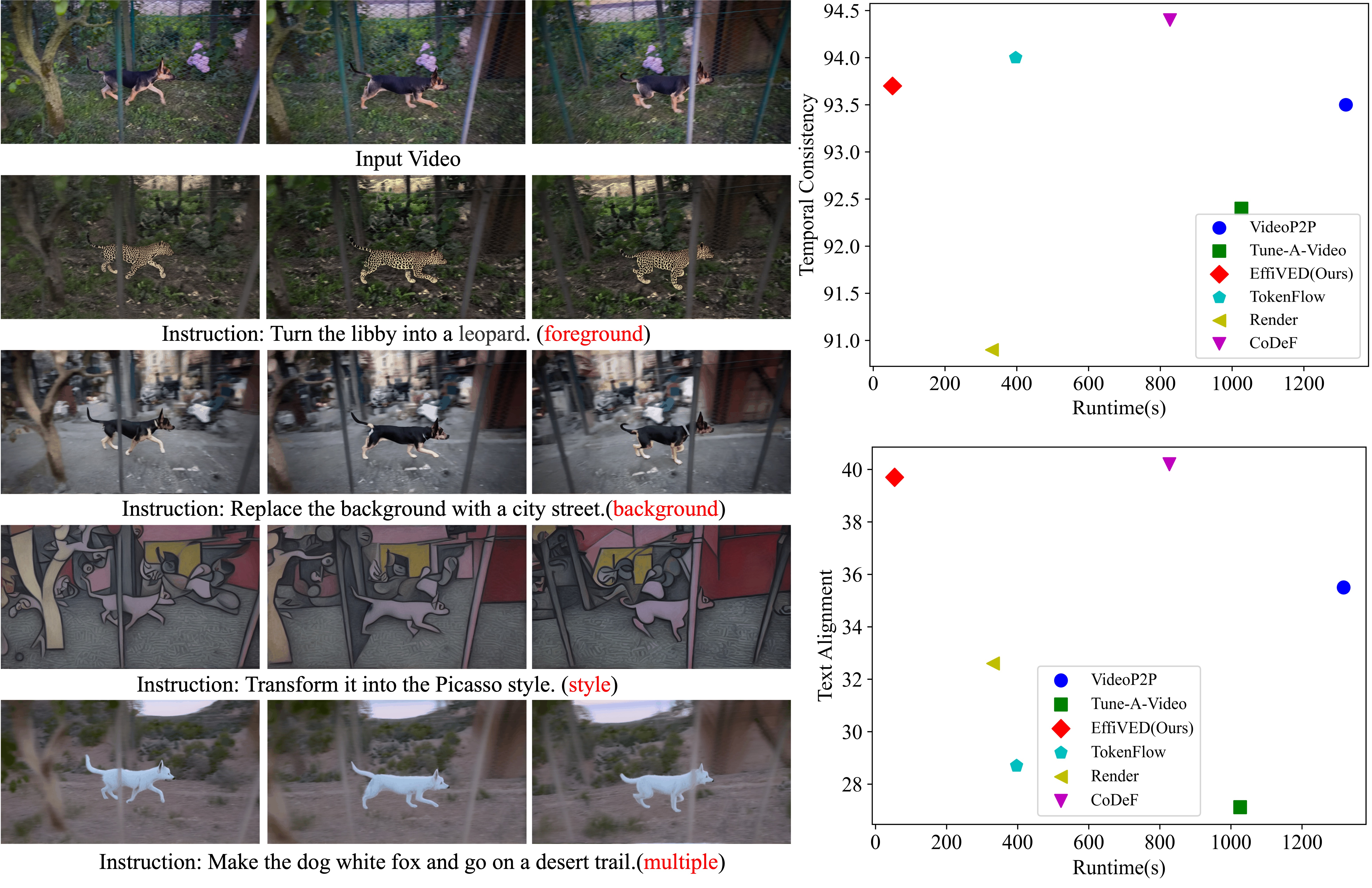}
    \caption{
        \textbf{On the left:} EffiVED offers users a versatile range of video editing capabilities, including modifications to objects, backgrounds and style transfer. \textbf{On the right:} Temporal Consistency \& Text Alignment \emph{vs}. Runtime(s) comparison on the TGVE dataset\cite{DBLP:journals/corr/abs-2401-07781}.  For runtime, all methods are evaluated by editing a 60-frame, 512p$\times$512p video using A100 GPUs with the official implementation. EffiVED achieves an impressive inference speed of 47 seconds, offering a 6 to 28 times speed boost compared to existing methods without compromising the quality of editing.
    }
    \label{fig:1}   
\end{figure*}

To address the aforementioned problem, we propose EffiVED, an open-domain video editing framework that seamlessly edits input videos according to human language instructions without necessitating any specific fine-tuning. EffiVED is a conditional diffusion model that employs a conditional 3D U-Net\cite{ronneberger2015u} architecture at its core. This design enables it to generate edited video content directly from the provided input video and associated textual instructions. 

To achieve direct video editing, we need a diverse training dataset that pairs input videos with instructions and edited results. This dataset must maintain temporal consistency and precisely execute the instructed changes. Unfortunately, no current readily available dataset or model meets these specific needs, and collecting real-world video-instruction pairs for training is laborious and costly due to the meticulous alignment required between visuals and text descriptions.

Inspired by Instruct-Pix2Pix\cite{DBLP:conf/cvpr/BrooksHE23} that integrates multiple foundational models\cite{DBLP:conf/nips/BrownMRSKDNSSAA20,rombach2022high} to create the image editing pairs, we have created a synthetic video dataset that pairs instructions with their before-and-after visual counterparts. To maximize the utilization of extensive instruction-based image editing datasets\cite{DBLP:conf/cvpr/BrooksHE23,DBLP:journals/corr/abs-2306-10012}, we apply augmentations such as random affine transformations to source and target images, simulating camera movements. This strategy enables us to seamlessly generate video triplets. These synthetic data instances provide rich visual context and enhance alignment between the visual and textual domains. 
To expand the variety of motion patterns, we also incorporate a large language model (LLM)\cite{DBLP:journals/corr/abs-2303-08774}  and the CoDeF technique specifically designed for real-world video content. Specially, we first leverage an example-driven approach to guide the LLM, empowering it to generate accurate descriptions that align with both videos and editing instructions. Subsequently, we train a CoDeF model to generate the corresponding edited counterparts, taking advantage of its demonstrated superior fidelity.

Employing video editing pairs from these methodologies, we create a robust training foundation for EffiVED. During training, we first extract video latents with a Variational Autoencoder (VAE) \cite{DBLP:journals/corr/KingmaW13}. These latents are then combined with their noisy versions and denoised back to the edited video latents. We separate classifier-free guidance into two distinct components: vision-conditioned and text-conditioned guidance. This separation allows each part to independently control the editing process based on visual or textual information. Consequently, it ensures generated videos more accurately adhere to both the original video content and the given text instructions.

In summary, our principal contributions are as follows:(i) we introduce two workflows for the collection of video editing pairs. This process effectively converts large-scale image editing datasets and open-world videos into a high-quality training set for video editing. With this workflow, we (ii) train a video editing model EffiVED that is capable of performing a broad range of editing tasks at an impressively fast speed, as evidenced in \cref{fig:1}. (iii) To investigate how our data collection approach impacts editing performance, we carry out several domain-specific experiments. These experiments explore strategies such as using data augmentations for image editing data, assessing video caption quality, and integrating multiple vision-language models in real videos. These may help overcome the data scarcity observed in the video editing domain.


\section{Related Work}

\subsection{Text-to-Video Generation}
Recent advancements in diffusion models have shown great promise in video generation \cite{nichol2021glide, saharia2022photorealistic, rombach2022high, peebles2023scalable, podell2023sdxl}. The Video Diffusion Model (VDM) \cite{ho2022video} pioneers this domain by adapting the image diffusion U-Net architecture \cite{ronneberger2015u} into a 3D structure for joint image and video training. Imagen Video \cite{ho2022imagen} employs a series of video diffusion models for high-resolution, temporally coherent video synthesis. Make-A-Video \cite{singer2022make} innovatively learns motion patterns from unlabeled video data, while Tune-A-Video \cite{wu2023tune} explores one-shot video generation by fine-tuning LDMs with a single text-video pair. Text2Video-Zero \cite{khachatryan2023text2video} tackles zero-shot video generation using pretrained LDMs without further training. ControlVideo \cite{zhang2023controlvideo} introduces a hierarchical sampler and memory-efficient framework to craft extended videos swiftly. 
Despite these innovations, capturing complex motion and camera dynamics remains challenging. VideoComposer \cite{wang2023videocomposer} and DragNUWA \cite{yin2023dragnuwa} propose motion trajectory-based control for video generation, yet they fall short in interactive animation with multiple objects. I2VGen-XL\cite{DBLP:journals/corr/abs-2311-04145} reduces the reliance on well-aligned text-video pairs by utilizing a single static image as the primary condition. AnimateAnything\cite{DBLP:journals/corr/abs-2311-12886} additionally incorporates motion masks and motion strength to precisely manipulate individual objects within an image, thus facilitating more user-centric interactive animation generation. Building upon the significant advancements in those above video generation models, our approach focuses on enhancing the fidelity of VDMs, particularly in the realm of video content translation.

\subsection{Text-Guidance Image Editing}
Text-Guidance Image Editing\cite{DBLP:conf/iccv/AbdalQW19, DBLP:conf/cvpr/AvrahamiLF22,DBLP:conf/eccv/CrowsonBKSHCR22,DBLP:journals/jmlr/HoSCFNS22} is a sophisticated process that modifies images according to precise text instructions. Early methods concentrated on adjusting the reverse diffusion process to maintain faithfulness. SDEdit\cite{DBLP:conf/iclr/MengHSSWZE22} pioneers as the first diffusion-based technique, applying varied noise levels to source images and generating edited results through diffusion sampling directed by edit prompts. Plug-and-Play\cite{DBLP:conf/cvpr/TumanyanGBD23} starts with DDIM inversion\cite{DBLP:conf/iclr/SongME21} to sample edited videos, copying selected visual features during diffusion. Prompt-to-Prompt(P2P) \cite{DBLP:conf/iclr/HertzMTAPC23} allows for a variety of edits by substituting the attention mechanisms in generated images with those from source images during the diffusion process. Other methods use optimization techniques for better editing. For instance, Imagic \cite{DBLP:conf/cvpr/KawarZLTCDMI23} employs textual inversion concepts from a specific source \cite{Gal2022image}, while Null-text Inversion\cite{DBLP:conf/cvpr/MokadyHAPC23} extends P2P to regulate cross-attention behavior by optimizing null text embeddings, thereby enabling more accurate reconstruction of the original image. However, these methods typically necessitate time-consuming because of per-image optimization. Recently, approaches like Instruct-Pix2Pix\cite{DBLP:conf/cvpr/BrooksHE23} have treated image editing as a supervised learning task, collecting paired synthetic data and fine-tuning text-to-image models\cite{DBLP:conf/cvpr/RombachBLEO22,DBLP:journals/corr/abs-2307-01952}. InstructDiffusion\cite{DBLP:conf/cvpr/BrooksHE23} consolidates multiple vision tasks under this framework. 

\subsection{Text-Guidance Video Editing}
Video editing is a substantially more challenging task~\cite{DBLP:journals/corr/abs-2307-10373,DBLP:conf/eccv/ZhuoWLW022,DBLP:conf/nips/Wang0TLCK19,DBLP:conf/cvpr/FuWGEW22} than image editing, particularly due to the need for frame-to-frame consistency and addressing temporal inconsistencies. Current approaches mainly focus on effectively steering large pre-trained models and can be broadly grouped into three categories: zero-shot, one-shot, and feature propagation techniques. Zero-shot approaches typically involve incorporating cross-frame attention or token flow mechanisms to ensure temporal coherence. For instance, Video-P2P\cite{DBLP:journals/corr/abs-2303-04761} employs a decoupled guidance strategy for real-world video editing tasks, utilizing cross-attention control. Similarly, Pix2Video\cite{DBLP:journals/corr/abs-2303-12688} and FateZero\cite{DBLP:conf/iccv/QiCZLWSC23} both propose distinct variations on replacing self-attention with cross-frame attention mechanisms. However, given that these models are not specifically trained on real-world videos, they tend to struggle with preserving temporal consistency in complex video scenarios. Moreover, the per-frame application of cross-attention comes with an additional computational cost.
One-shot methods utilize fine-tuned diffusion models on a given video, which then sample new content conditioned upon specific edit prompts. For instance, Tune-A-Video\cite{DBLP:conf/iccv/WuGWLGSHSQS23} and SimDA\cite{DBLP:journals/corr/abs-2308-09710} generate videos based on textual prompts through the process of one-shot fine-tuning of pre-trained stable diffusion models. Dreamix\cite{DBLP:journals/corr/abs-2302-01329} introduces a groundbreaking mixed fine-tuning model that notably enhances the quality of motion edits. Meanwhile, AnimateDiff\cite{DBLP:journals/corr/abs-2307-04725} leverages customized model weights to incorporate conditional content, achieved either by using LoRA\cite{DBLP:conf/iclr/HuSWALWWC22} or DreamBooth\cite{DBLP:conf/cvpr/RuizLJPRA23}. 
Recently, Neural Layered Atlases\cite{DBLP:journals/tog/KastenOWD21} have gained popularity for their ability to decompose input into consistent layered representations. Text2Live\cite{DBLP:conf/eccv/Bar-TalOFKD22} augments atlases with extra edit layers and trains a dedicated generator for them. StableVideo\cite{DBLP:journals/corr/abs-2308-09592} improves temporal consistency through an atlas aggregation network. CoDeF\cite{DBLP:journals/corr/abs-2308-07926} estimates canonical images and temporal deformation fields using optical flow from source videos. Despite its ability to produce high-quality edited videos with similar motion patterns, this approach inherently involves repetitive and laborious fine-tuning for each individual video input, which restricts its scalability across broader applications. In contrast, our method leverages VDM to execute video editing tasks directly by using our synthetic dataset, thus ensuring temporal coherence without additional fine-tuning.

\section{Method}
In this section, we first introduce the preliminaries of the LDM in ~\cref{3.1}. Next, we present the workflow of the dataset collection used for training in ~\cref{3.2}. Finally, we describe the architecture of EffiVED in ~\cref{3.3}.

\subsection{Preliminaries of LDM}
\label{3.1}
In this section, we present the foundational concepts of Latent Diffusion Models (LDM) \cite{rombach2022high}. Given an image sample $x_0 \in \mathbb{R}^{3 \times H \times W}$, LDM initially employs a pre-trained VAE to compress $x_0$ into a lower-resolution latent representation $z_0 \in \mathbb{R}^{c \times h \times w}$. The forward process in LDM can be depicted as a Markov chain that progressively adds Gaussian noise to the latent representation step by step:
\begin{equation}
    q(z_t|z_{t-1}) = \mathcal{N}(z_t; \sqrt{1-\beta_t}z_{t-1}, \beta_tI),
\end{equation}
where $t=1,...,T$, $T$ is the total number of timesteps. $\beta_t$ denotes a coefficient that controls the intensity of the noise in step $t$.
The iterative noise adding can be simplified as:

\begin{equation}
    z_t=\sqrt{\bar{\alpha}_t}z_0 + \sqrt{1 - \bar{\alpha}_t}\epsilon, \quad\epsilon \sim \mathcal{N}(0, I),
    \label{eq:add_noise}
\end{equation}
where $\bar{\alpha}_t = \prod_{i=1}^t(1-\beta_t)$. During training, the LDM learns to approximate the latent space distribution of authentic data by predicting the noise $\epsilon$ applied to $z_t$, thereby reducing computational complexity compared to conventional diffusion models. The objective function for this learning process can be formulated as:
\begin{equation}
    l_\epsilon = ||\epsilon - \epsilon_\theta(z_t, t, c)||^2_2,
    \label{eq:training_objective}
\end{equation}
where $\epsilon_\theta(\cdot)$ denotes the noise prediction function of diffusion models. LDM simplifies flexible control of generation by converting user conditions $c$ with a domain-specific encoder into an intermediate format, which is then incorporated into the UNet via cross-attention.

Video diffusion models like~\cite{ho2022video, singer2022make, wang2023videocomposer} build upon image LDMs by incorporating a 3D U-Net structure, thereby allowing them to adeptly process video content. This 3D U-Net adds temporal convolutions after each spatial one and follows every spatial attention layer with a temporal attention block. To ensure it retains the capacity to generate from image data, the 3D U-Net is trained simultaneously using both image and video datasets.
\subsection{Training Data Construction}
\label{3.2}
Given source video $V_s$ and instruction $c$, we aim to generate target video $V_t$ without per-video fine-tuning. We achieve this by assembling video triplets of various editing situations for training, then conditionally translate $V_s$ into $V_t$ based on $c$. This section discusses our strategy for collecting datasets.

\begin{figure*}[h]
    \vspace{-3mm}
    \centering
    \includegraphics[width=0.94\textwidth]{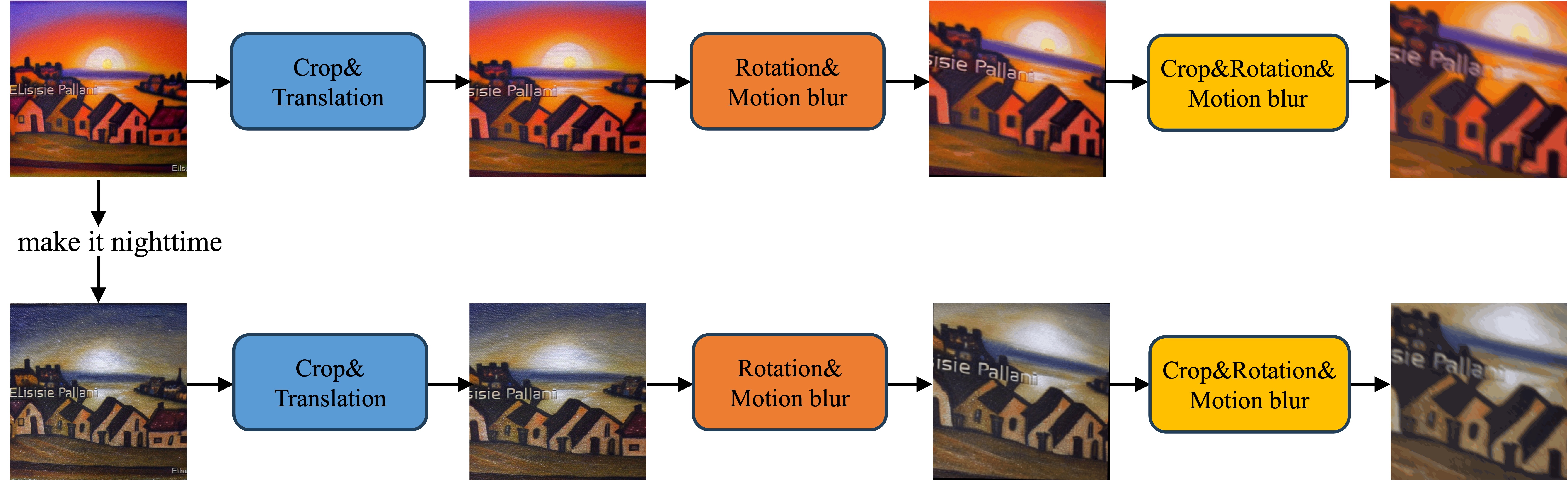}
    \caption{
     An example of generating training data from image editing dataset.  Given pairs of original and edited images, we randomly select and apply a set of affine transformations (e.g., rotation, crop, translation, or shearing) to both images. This approach generates a sequence of frames that simulate camera movement for each image. 
    }
    \label{fig:img_data}   
    \vspace{-2.5mm}
\end{figure*}

\noindent \textbf{Generating with image editing datasets.} InstructPix2Pix~\cite{DBLP:conf/cvpr/BrooksHE23} leverages LLM to generate a large dataset of image editing examples. Given the relative scarcity of videos compared to images, we capitalize on the following insight to synthesize videos that simulate camera movements using these image samples. In particular, if a frame $I_{t}$ differs from previous frame $I_{t-1}$ only in the camera position, then the edited frame $E_{t}$ and $E_{t-1}$ should only be different in the camera position as well.

Given a pair of images, namely the original and the edited, represented as ($I$, $E$), we randomly select and apply a set of affine transformations or random crops $\{F\}_{t=1}^{T} $ to both images. As shown in \cref{fig:img_data}, this process results in the generation of the pseudo video sequence with a length of $T$:
\begin{equation}
    V_{I} =  \{F(I)\}_{t=1}^{T},   V_{E} =  \{F(E)\}_{t=1}^{T}
\end{equation}

We set random rotations degrees to < 5 , random translation to [$-$0.05,0.05], random scaling factor to [0.95,1.05], and random shear degrees to $[-5^\circ,5^\circ ]$ on both axis. For the random resized crop, we scale the original image to 288 pix, and randomly crop a square image with 256 pix.
\begin{figure*}[!h]
    \vspace{-5.5mm}
    \centering
    \includegraphics[width=0.95\textwidth]{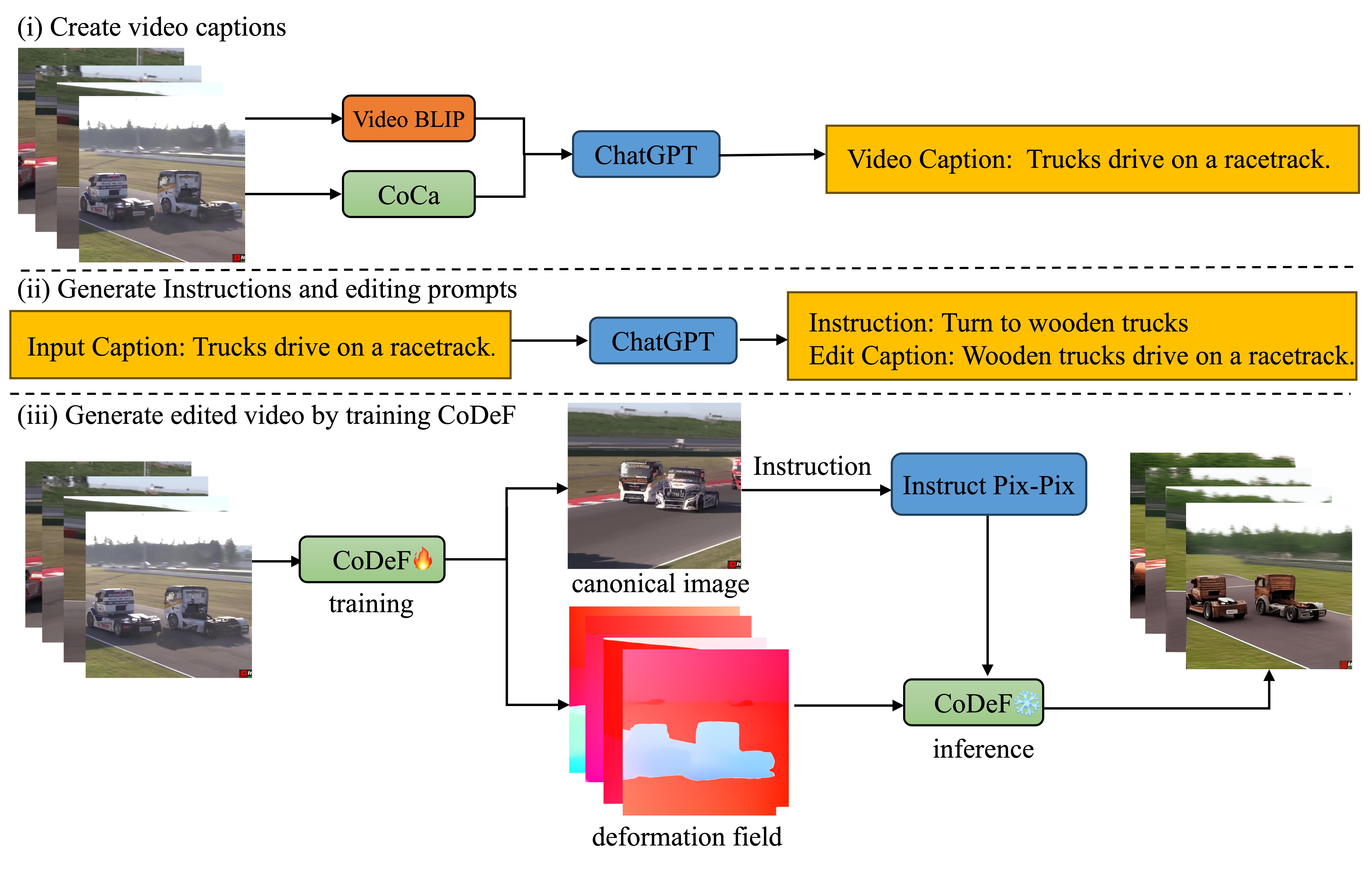}
    \caption{
    An overview of generating training data with open-world videos. (i) First, we leverage CoCa and VideoBLIP to extract caption from both keyframes and the entire video content, which are then synthesized into a comprehensive summary by ChatGPT. (ii) Next, we utilize ChatGPT to generate editing instruction and edited caption by providing manually examples. (iii) Finally, the generated instruction and the original video feed an individual CoDeF model to produce edited video.
    }
    \vspace{-2mm}
    \label{fig:openworld}  
    \vspace{-3mm}
\end{figure*}

\noindent \textbf{Generating with open-world videos.} While generating from image editing datasets can indeed offer strong alignment between text and visual content, it inherently lacks the natural motion dynamics found in real-world videos. In this section, we will illustrate a pipeline that leverages model composition to generate instructions and corresponding edited counterparts from open-world videos. This is achieved by integrating multiple fundamental models including ChatGPT\cite{DBLP:journals/corr/abs-2303-08774}, CoCa\cite{DBLP:journals/tmlr/YuWVYSW22}, V-BLIP\cite{VBLIP} and CoDeF\cite{DBLP:journals/corr/abs-2308-07926}.

As shown in \cref{fig:openworld}, the first step is to generate video captions for open-world video clips. Motivated by SVD\cite{DBLP:journals/corr/abs-2311-15127}, we follow a multi-step approach for generating an accurate and comprehensive video caption. In a nutshell, we uniformly extract four key-frames from the video and use CoCa to generate captions for each. Subsequently, V-BLIP generates a full-video caption by interpreting context across these frames. Finally, ChatGPT consolidates all five captions into a concise summary that captures the essence of the entire video content.

Based on the generated video captions, we utilize ChatGPT to craft instructions and edited captions accordingly. This AI tool can ingeniously derive contextually logical instructions and captions based on a minimal number of manual examples. For more instances showcasing ChatGPT's in-context learning capabilities, please refer to the supplementary materials.


Finally, we employ a modified CoDeF to create edited videos that follow given instructions while preserving subtle motions. This method involves fine-tuning per video to construct a canonical content field that consolidates the static contents  throughout the entire video and a temporal deformation field which records the transformations from the canonical image to every frame over time. These two fields are model by dynamic Neural Radiance Fields (NeRFs)\cite{DBLP:conf/iccv/ParkSBBGSM21}. To enhance the modeling of complex motion, we augment dynamic NeRFs with an increased hash table size to accommodate more grids at varying resolutions. This allows us to capture more high-frequency details, thereby improving the quality of edited videos. Upon training, CoDeF saves the canonical image and its corresponding deformation field for each video. When it comes to generating the edited video, an image-editing model such as InstructPix2Pix\cite{DBLP:conf/cvpr/BrooksHE23} is applied to the canonical image based on the modified caption to produce the edited canonical image. Then, this edited canonical image is propagated to all frames using the saved deformation fields, thereby creating the final edited video with maintained consistency and structure. 

Through these automated processes, we can systematically generate multiple sets of input videos, instructions, and their corresponding edited videos. These triplets serve as the valuable training data set for our EffiVED model, enhancing its capability to understand and execute editing instructions effectively on different videos.
\subsection{EffiVED Model}
\label{3.3}

\begin{figure*}
    \centering
    \includegraphics[width=\textwidth]{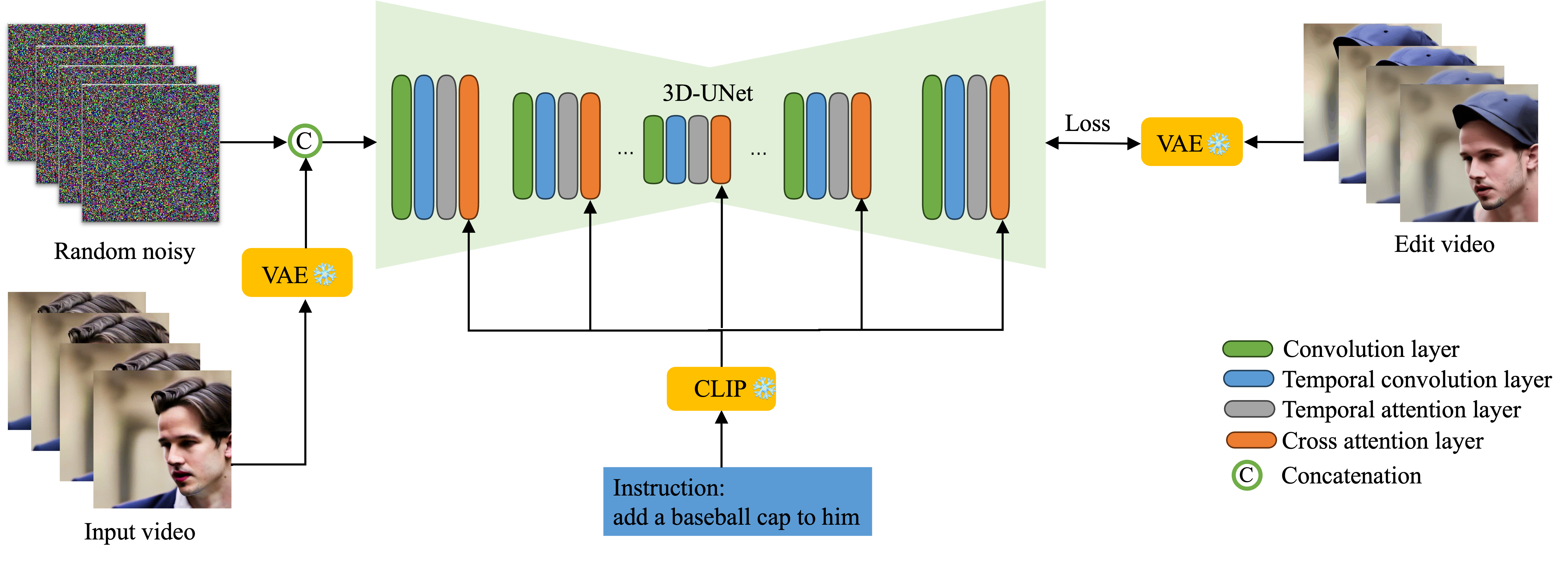}
    \caption{
        Overview of our training pipeline. We adopt the widely used 3D U-Net based video diffusion model~\cite{wang2023videocomposer} for video editing. To enable vision conditioning, we augment the 3D-UNet's input by appending extra channels to its initial convolutional layers. The input is essentially a channel-wise concatenation of video latents and noise.
    }
    \vspace{-5mm}
    \label{fig:pipeline}   
\end{figure*}

\noindent \textbf{Training pipeline of EffiVED}. 
The detailed training pipeline for EffiVED is depicted in \cref{fig:pipeline}. Given the input video $V_{I}$ and the editing video $V_{E}$, we utilize a pre-trained VAE encoder to transform these videos into their latent space representations $x_{I}$ and $x_{E}$. Then, the sampled noisy latent $\epsilon$ is concatenated with the latent representation of the input video along the channel dimension and fed into the 3D U-Net. Throughout this process, the text embedding serves to guide the cross-attention mechanism. Here, both the input video $x_{I}$ and the provided instruction $c$ act as conditions that control the denoising process, thereby steering the translation towards the editing video $x_{E}$. We minimize the following latent diffusion objective:

\begin{equation}
    \mathbb{E}_{x_I,x_E,\epsilon \sim \mathcal{N}(0,1),t}\left [ \left | \epsilon - \epsilon_{\theta}(\left [ x_{I},x_{E} \right ],c,t ) \right |_{2}^{2}  \right ]  ,
\end{equation}
where $t$ is the denoising step, the optimization objective centers on accurately estimating the introduced noise while concurrently maintaining temporal coherence among neighboring frames.

\noindent \textbf{Decoupled Classifier-Free Guidance}. Given that our method accepts both instructions and videos as input, it needs to balance the probability mass of these two streams when guiding the output. Inspired by \cite{wang2023videocomposer} that skillfully combines multiple prompts for accurate score estimates, we randomly choose either the input video latent $x_{I}$ or the text prompt $c$ as the only input. During inference,  the predicted noise at step $t$ can be computed as:
\begin{equation}
\begin{split}
\widetilde{\epsilon}(x_{I},c,t) =  \widetilde{\epsilon}(\phi,\phi ,t) + \lambda _{1} (\widetilde{\epsilon}(x_{I},c ,t) -\widetilde{\epsilon}(\phi,c ,t) ) \\  
+ \lambda _{2} (\widetilde{\epsilon}(x_{I},c ,t) -\widetilde{\epsilon}(x_{I},\phi  ,t) ),
\end{split}
\end{equation}
where $\lambda _{1}$ and $\lambda _{2}$ denote the text and vision guidance scales, respectively. A higher value of either $\lambda _{1}$ or $\lambda _{2}$ will exert a stronger influence on directing the edited video to adhere more closely to the corresponding condition.
\section{Experiments}
\subsection{Experiments Setup}
\textbf{Dataset}. For our evaluation, we use the Text-Guided Video Editing (TGVE)\cite{DBLP:journals/corr/abs-2401-07781} competition dataset that contains 76 videos. Every video in the dataset comes with one original prompt that describes the video and four prompts that suggest different edits for each video. Three editing prompts pertain to modifications in style, background, or object within the video. Additionally, a multiple editing prompt is provided that may incorporate all three types of edits simultaneously.

\noindent \textbf{Metrics}. Following previous works, we assess the edited videos based on two main aspects: text-video alignment and video quality. The text-video alignment measures how closely the edited video adheres to the provided text instructions. This is calculated by averaging the text-image similarity in the embedding space of CLIP image models across all frames in the video. We also utilize PickScore\cite{DBLP:journals/corr/abs-2305-01569} to gauge whether the structural integrity of the original video has been successfully maintained after editing. Regarding video quality, we examine frame consistency, which is estimated by calculating the average cosine similarity between CLIP image embeddings for each pair of edited frames. This helps us understand how well the visual coherence is maintained.

\noindent \textbf{Implementation Details.} We first collect 131k training clips, each containing eight frames, from InstructPix-Pix\cite{DBLP:conf/cvpr/BrooksHE23} and MagicBrush\cite{DBLP:journals/corr/abs-2306-10012} using the proposed method. For generating instructions and edit prompts from open-world videos, we follow the approach outlined in InstructPix-Pix to guide ChatGPT. We utilize the official repositories of CoDeF and only increase the hash table size for better reconstruction of high-frequency details. During this process, we collect 24k training clips. The EffiVED model is initialized with Modelscope T2V\cite{DBLP:journals/corr/abs-2308-06571}. Our training strategy employs a two-stage methodology. Initially, the model undergoes training for 30k iterations on a dataset of 131k clips with a batch size of 4, enabling it to effectively align with editing instructions and adeptly manage various editing tasks. In the subsequent phase, refinement continues by training the model on open-world videos for an additional 10k iterations, which thereby enhances its temporal consistency. To further optimize performance, during this latter stage, we implement multi-frame rate sampling, extracting training clips of 8-frames from a variety of frame rates (e.g., 4, 6, and 8).

\subsection{Evaluation Results}
In \cref{tab1}, we present the quantitative results of several mainstream video editing models. It is evident that the majority of these methods rely on one-shot tuning or inversion optimization, significantly impeding practical application. CoDeF attains the highest level of editing quality due to its superior reconstruction. By training on our dataset, our method sustains a comparable performance in terms of fidelity and temporal consistency. Regarding runtime efficiency, it is worth noting that our method does not require any finetuning or inversion, and as a result, it operates around 6 times faster than Render-A-Video\cite{DBLP:conf/siggrapha/YangZLL23} and an impressive 20 times faster than CoDeF. This substantial difference in speed effectively showcases the high level of efficiency inherent in our proposed method.
\begin{table}[h]
\vspace{-3mm}
\centering
\renewcommand{\arraystretch}{0.9}
\begin{tabular}{c|c|c|c|c}
\hline
Method          & Text Alignment & Frame Consistency & PickScore & Runtime(s) \\ \hline
Text2Video-Zero\cite{khachatryan2023text2video} & 25.9          & 92.1             & 19.9     &    -    \\
Render-A-Video\cite{DBLP:conf/siggrapha/YangZLL23}  & 32.6           & 90.9              & 19.6     &    294     \\ 
Vid2Vid-Zero\cite{DBLP:journals/corr/abs-2303-17599}    & 40.0          & 92.6              & 20.4     &     -    \\ 
Video-P2P\cite{DBLP:journals/corr/abs-2303-04761}       & 35.5           & 93.5              & 20.1     &   1385     \\ 
Tune-A-Video\cite{DBLP:conf/iccv/WuGWLGSHSQS23}    & 27.1          & 92.4              & 20.3     &   1026      \\ 
TokenFlow\cite{DBLP:journals/corr/abs-2307-10373} & 28.7 & 94.0 & 20.5 & 394\\
CoDeF\cite{DBLP:journals/corr/abs-2308-07926}           & \textbf{40.2}           & \textbf{94.2}              & \textbf{20.8}     &    827     \\ 
\textbf{EffiVED}         & 39.7           & 93.7              & 20.6     &    \textbf{47}     \\  \hline
\end{tabular}
\caption{Comparision of editing quality and runtime on TGVE dataset. We measured the running time of 60 frames. 
}
\label{tab1}
\vspace{-6mm}
\end{table}

\begin{figure*}[h]
    \centering
    \includegraphics[width=\textwidth]{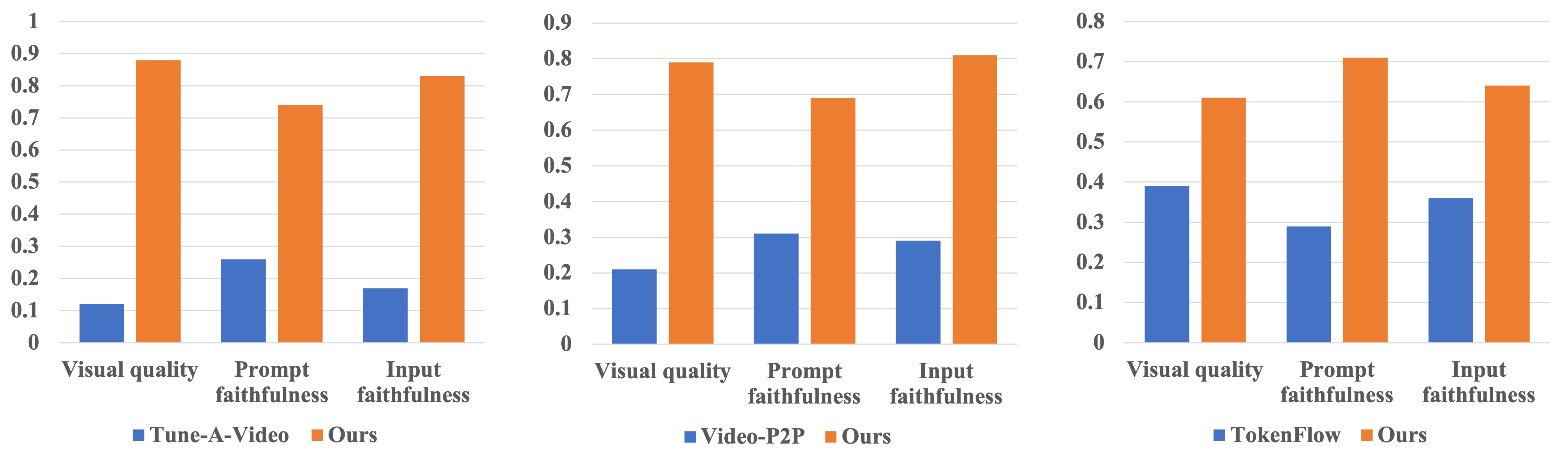}
    \caption{
        A/B Comparison with current methods. Our method not only effectively aligns edited videos with instructions but also consistently preserves the video's structure.
    }
    \label{fig:user}   
    \vspace{-5mm}
\end{figure*}

We also conduct an A/B testing to evaluate the superiority of our method against three techniques: Tune-A-Video, Video-P2P, and TokenFlow. We engage five human evaluators to assess the quality of a sample of 100 generated videos. The evaluation is based on three criteria. The first criterion assesses visual quality, encompassing clarity and absence of flicker. The second metric evaluates prompt faithfulness, determining how well the output video reflects and executes the content in the editing instructions. Lastly, input faithfulness checks if the edited videos retain the original video's structure. As shown in \cref{fig:user}, in terms of visual quality, our method are notably better than Tuna-A-Video and Video-P2P. This improvement can be attributed to the synthesis video data for training, our model learn better to keep the temporal consistency and maintains the structural integrity with the condition of the origin video latent. For faithfulness, our method are notably better than all compared methods, showing that the efficiency of the fully use of the image editing dataset and the generating captions can better align the origin videos. 

\begin{figure*}[!h]
    \centering
    \includegraphics[width=0.9\textwidth]{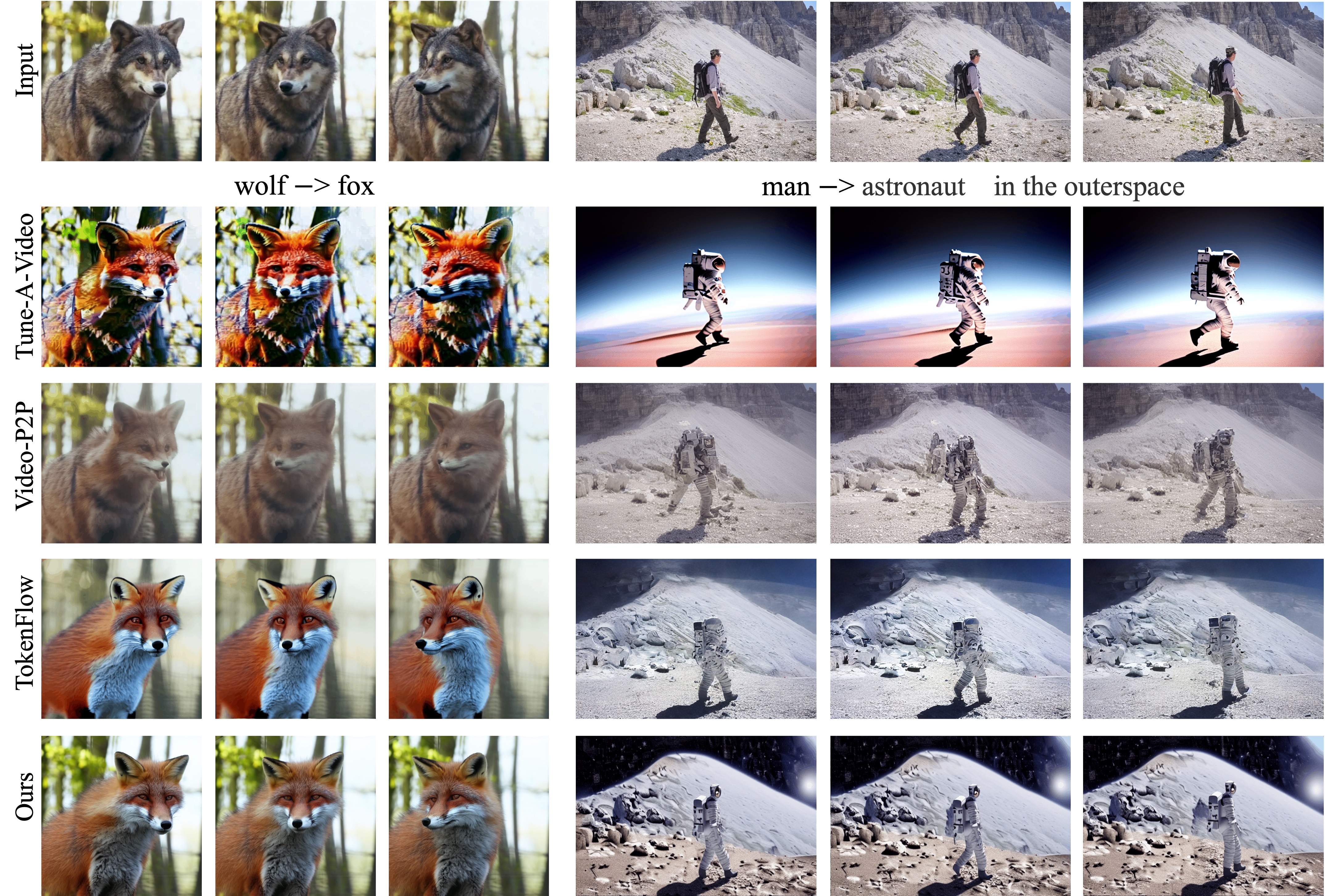}
    \caption{
        Qualitative Comparison. EffiVED outperforms baseline methods by demonstrating superior consistency and a heightened adherence to the provided instructions.}
    \label{fig:compare}   
\end{figure*}

\begin{figure*}[!h]
    \centering
    \includegraphics[width=0.89\textwidth]{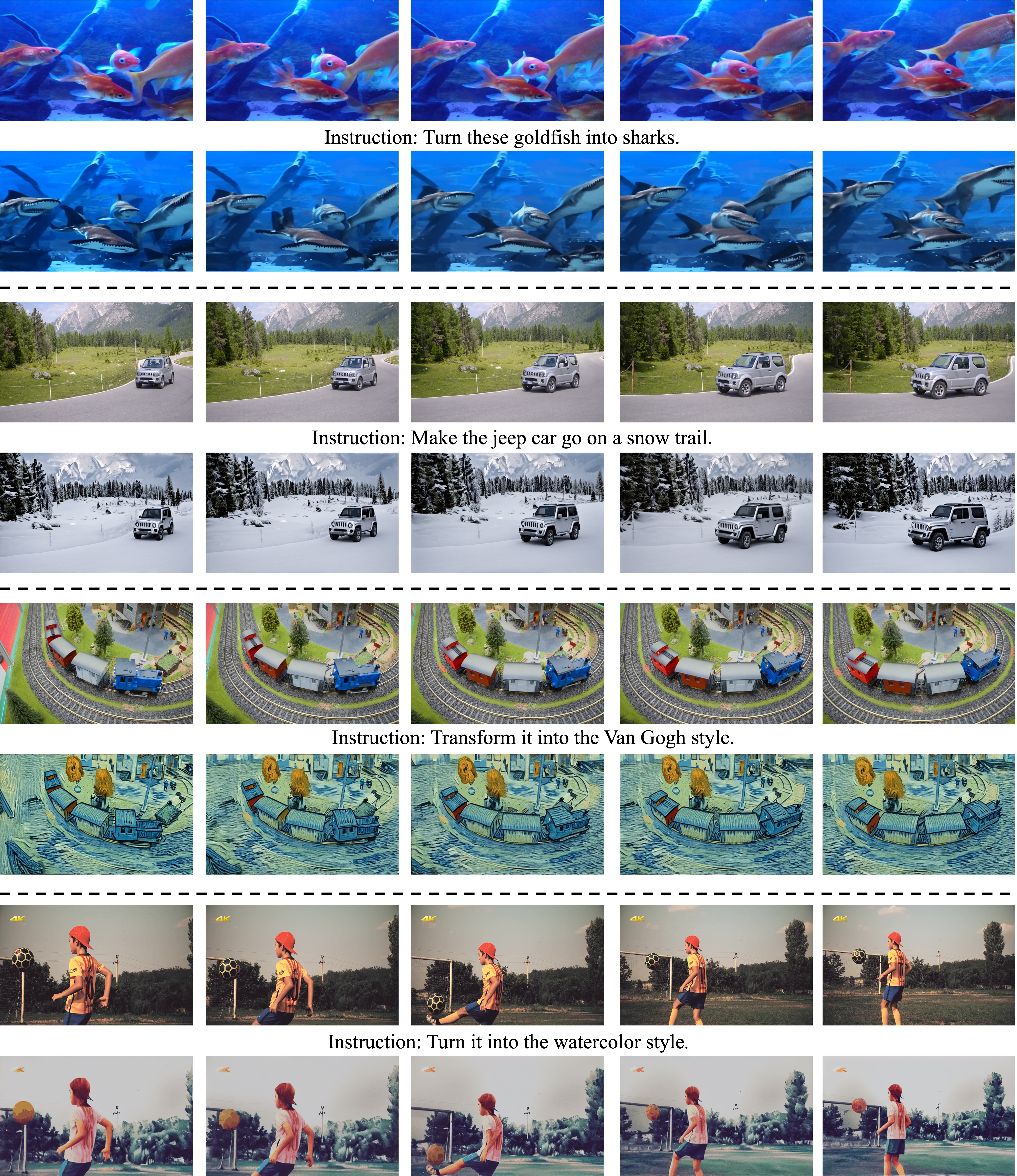}
    \caption{
        Qualitative results of EffiVED. EffiVED offers flexible editing capabilities, including changes to backgrounds, objects, and styles.
    }
    \label{fig:quantative}  
    \vspace{-6mm}
\end{figure*}

Moreover, we present a qualitative comparison in \cref{fig:compare} with these methods. Note that Video-P2P, untrained on video data and adjusting cross-attention solely between prompts, often fails to preserve object consistency across frames (as evidenced by the astronaut and fox). Tune-A-Video struggles with generating coherent motion when handling multiple edits, as seen in the astronaut's actions. TokenFlow, while adept at maintaining structural continuity, suffers from a 'feature-level smoothing' drawback, causing blurring in static areas, evident in the fox where backgrounds are blurred compared to the original versions. In contrast, our method stands out by preserving global consistency throughout the edited videos, ensuring alignment with the given text prompt, and upholding high editing quality—all achieved within a shorter time compared to previous methods. In \cref{fig:quantative}, we show more editing cases including stylization(e.g. watercolor style, Van Gogh style), background(e.g replacing the highway with snow trail), multiple objects (e.g. transforming several goldfish into sharks).

\subsection{Ablation Studies}

In this section, we perform several ablation studies to validate the effectiveness of our data collection strategies. All results are evaluated on TGVE datasets.

\noindent \textbf{The source and scale of training data.} 
 We train EffiVED with separate datasets: the 131k synthetic pairs from an image editing dataset and the 23k samples from open-world videos. We also conduct training with mixed datasets of varying sizes from both sources to analyze performance fluctuations. The results of these experiments are presented in \cref{tab2}.
\begin{table}[h]
\vspace{-3mm}
\centering
\begin{tabular}{c|c|c|c}
\hline
Training Data          & Text Alignment & Frame Consistency & PickScore  \\ \hline
131k from image editing & 37.7          & 91.6             & 20.1        \\
23k from open-world video  & 35.6           & 92.3              & 19.4          \\ \hline
Mixed data(20k) &         32.6   &    68.7    &  18.9  \\ 
Mixed data(50k) &         35.7   &    81.3    &   19.7 \\
Mixed data(80k) &         37.9   &    89.6    &   20.3 \\
Mixed data(120k)&        38.8    &    92.9    &    20.4                \\    
Mixed data(all)&       39.1     &   93.7     &     20.6             \\     \hline
\end{tabular}
\caption{Ablation studies on the source and scale of training data. The mixed data refers to the combination of datasets from both sources, and we adopt a sampling ratio of 5:1 to collect the sub-dataset.}
\label{tab2}
\vspace{-5mm}
\end{table}

Training solely on image editing pairs yields good alignment with instructions but may suffer from temporal inconsistency due to unrealistic motion simulation. Open-world video data, while enhancing temporal coherence, can result in a less diverse and potentially less instruction-sensitive dataset. EffiVED effectively strikes a balance by leveraging both datasets, thereby ensuring accurate text-to-video alignment and maintaining robust temporal consistency. In terms of the training data volume, we observe a significant enhancement in performance as the size increases from 20k to 80k. This considerable improvement substantiates that our dataset indeed boosts the editing capabilities of the pretrained video diffusion model effectively. Upon further expansion of the dataset, the  performance improvement diminishes and eventually plateaus. Based on these empirical observations, we conclude that a total of 154k video samples is sufficient for attaining competitive results within our training process.

\begin{figure*}[]
    \centering
    \includegraphics[width=1.0\textwidth]{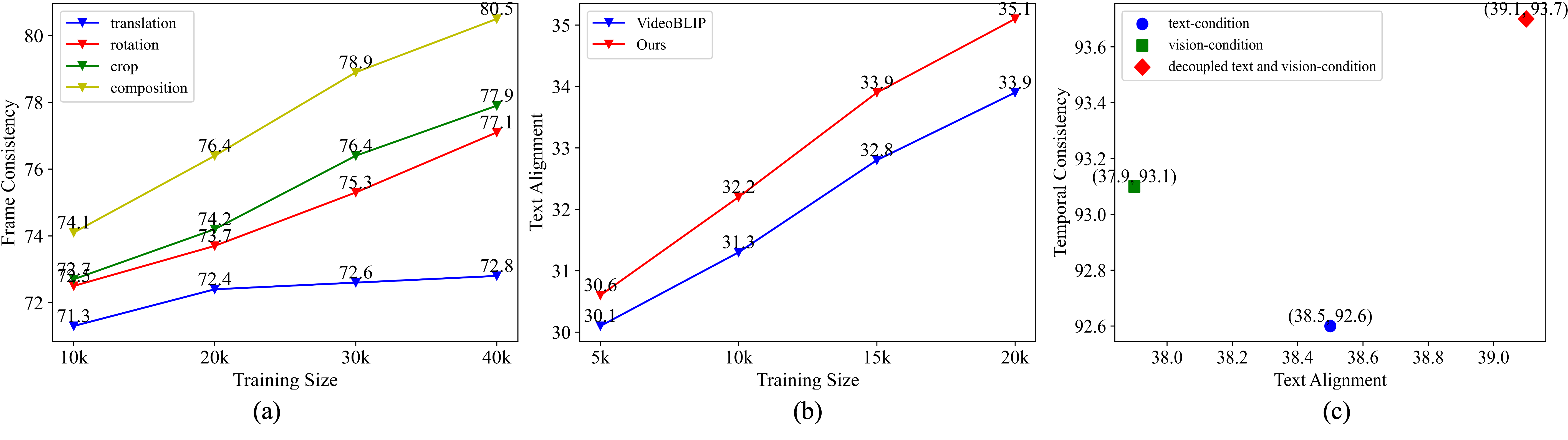}
    \caption{
    (a) The Frame consistency \emph{vs} training size of different augmentation strategies. (b) The Text alignment \emph{vs} training size of different captions generation methods. (c) The effectiveness of different classifier-free guidance strategies.
    }
    \label{fig:ab_compare}   
    \vspace{-3mm}
\end{figure*}

\noindent \textbf{Augmentation strategies.} To validate the effectiveness of different augmentation techniques, we randomly select identical edited pairs and apply various augmentations to them. The results are depicted in \cref{fig:ab_compare}(a). It's evident that applying only translation achieves convergence after using 30k training samples; however, it fails to yield a satisfactory temporal consistency. In contrast, rotation and random cropping significantly improve performance as the training data volume increases. We attribute this to the fact that random cropping more closely mimics real-world camera movements than mere translation. Additionally, rotation can simulate some degree of rigid motion. When these augmentations are combined, we demonstrates its excellence by achieving the best consistency.

\noindent \textbf{Caption quality.} We also conduct ablation studies on the quality of the generating captions from open-world videos. We utilize VideoBLIP as the sole baseline to generate the captions, and the comparative results are presented in \cref{fig:ab_compare} (b). Our method integrates key-frame captions from Coca and video captions from VideoBLIP, which describe spatial and temporal aspects respectively. It is apparent that our approach achieves an average of approximately 2$\%$ better text alignment compared to the baseline, and this difference widens as the training data size increases. More caption generation results can be found in the supplementary materials.

\noindent \textbf{Classifier-free guidance strategy.} We evaluate the performance of various classifier-free guidance strategies. As depicted in \cref{fig:ab_compare}(c), setting only text conditions with null embeddings for classifier-free guidance achieves a relatively higher text alignment score of 38.5$\%$, but this method struggles to maintain structural consistency in edited videos. In contrast, when only vision conditions are set with null embeddings, it fails to deliver high-fidelity results, scoring just 37.9$\%$ in text alignment. By decoupling the classifier-free guidance into separate text and vision guidance components, we observe significant improvements. This modification has led to an average improvement of 1$\%$ across both text-alignment accuracy and temporal consistency metrics compared to utilizing either text or vision guidance independently.


\section{Conclusion}

This paper introduces EffiVED, an efficient instruction-based video editing technique. We develop two innovative strategies to convert existing image editing datasets and real-world videos into a vast collection of video editing data. Through training on this amassed dataset, EffiVED can edit open-world videos directly without requiring per-video fine-tuning or inversion. Moreover, we integrate classifier-free guidance into both the video and text conditions. Experimental results demonstrate that our method significantly speeds up the editing process by approximately 6 to 28 times compared to current methods while maintaining a competitive level of editing quality. We hope our insights can help mitigate the challenge of data scarcity in the video editing.

\clearpage  

%
%
\bibliographystyle{splncs04}
\bibliography{egbib}

\begin{thebibliography}{10}
\providecommand{\url}[1]{\texttt{#1}}
\providecommand{\urlprefix}{URL }
\providecommand{\doi}[1]{https://doi.org/#1}

\bibitem{DBLP:conf/iccv/AbdalQW19}
Abdal, R., Qin, Y., Wonka, P.: Image2stylegan: How to embed images into the stylegan latent space? In: 2019 {IEEE/CVF} International Conference on Computer Vision. pp. 4431--4440 (2019)

\bibitem{DBLP:conf/cvpr/AvrahamiLF22}
Avrahami, O., Lischinski, D., Fried, O.: Blended diffusion for text-driven editing of natural images. In: {IEEE/CVF} Conference on Computer Vision and Pattern Recognition. pp. 18187--18197 (2022)

\bibitem{DBLP:conf/eccv/Bar-TalOFKD22}
Bar{-}Tal, O., Ofri{-}Amar, D., Fridman, R., Kasten, Y., Dekel, T.: Text2live: Text-driven layered image and video editing. In: Avidan, S., Brostow, G.J., Ciss{\'{e}}, M., Farinella, G.M., Hassner, T. (eds.) Computer Vision - {ECCV} 2022 - 17th European Conference, Tel Aviv. Lecture Notes in Computer Science, vol. 13675, pp. 707--723 (2022)

\bibitem{DBLP:journals/corr/abs-2311-15127}
Blattmann, A., Dockhorn, T., Kulal, S., Mendelevitch, D., Kilian, M., Lorenz, D., Levi, Y., English, Z., Voleti, V., Letts, A., Jampani, V., Rombach, R.: Stable video diffusion: Scaling latent video diffusion models to large datasets. arxiv  \textbf{abs/2311.15127} (2023)

\bibitem{DBLP:conf/cvpr/BrooksHE23}
Brooks, T., Holynski, A., Efros, A.A.: Instructpix2pix: Learning to follow image editing instructions. In: {IEEE/CVF} Conference on Computer Vision and Pattern Recognition. pp. 18392--18402 (2023)

\bibitem{DBLP:conf/nips/BrownMRSKDNSSAA20}
Brown, T.B., Mann, B., Ryder, N., Subbiah, M., Kaplan, J., Dhariwal, P., Neelakantan, A., Shyam, P., Sastry, G., Askell, A., Agarwal, S., Herbert{-}Voss, A., Krueger, G., Henighan, T., Child, R., Ramesh, A., Ziegler, D.M., Wu, J., Winter, C., Hesse, C., Chen, M., Sigler, E., Litwin, M., Gray, S., Chess, B., Clark, J., Berner, C., McCandlish, S., Radford, A., Sutskever, I., Amodei, D.: Language models are few-shot learners. In: Advances in Neural Information Processing Systems 33: Annual Conference on Neural Information Processing Systems 2020, NeurIPS 2020 (2020)

\bibitem{DBLP:journals/corr/abs-2303-12688}
Ceylan, D., Huang, C.P., Mitra, N.J.: Pix2video: Video editing using image diffusion. CoRR  \textbf{abs/2303.12688} (2023)

\bibitem{DBLP:journals/corr/abs-2308-09592}
Chai, W., Guo, X., Wang, G., Lu, Y.: Stablevideo: Text-driven consistency-aware diffusion video editing. CoRR  \textbf{abs/2308.09592} (2023)

\bibitem{DBLP:conf/eccv/CrowsonBKSHCR22}
Crowson, K., Biderman, S., Kornis, D., Stander, D., Hallahan, E., Castricato, L., Raff, E.: {VQGAN-CLIP:} open domain image generation and editing with natural language guidance. In: Avidan, S., Brostow, G.J., Ciss{\'{e}}, M., Farinella, G.M., Hassner, T. (eds.) Computer Vision - {ECCV} 2022 - 17th European Conference, Tel Aviv. Lecture Notes in Computer Science, vol. 13697, pp. 88--105 (2022)

\bibitem{DBLP:journals/corr/abs-2311-12886}
Dai, Z., Zhang, Z., Yao, Y., Qiu, B., Zhu, S., Qin, L., Wang, W.: Fine-grained open domain image animation with motion guidance. arxiv  (2023). \doi{10.48550/ARXIV.2311.12886}, \url{https://doi.org/10.48550/arXiv.2311.12886}

\bibitem{DBLP:conf/cvpr/FuWGEW22}
Fu, T., Wang, X.E., Grafton, S.T., Eckstein, M.P., Wang, W.Y.: M\({}^{\mbox{3}}\)l: Language-based video editing via multi-modal multi-level transformers. In: {IEEE/CVF} Conference on Computer Vision and Pattern Recognition. pp. 10503--10512 (2022)

\bibitem{DBLP:journals/corr/abs-2307-10373}
Geyer, M., Bar{-}Tal, O., Bagon, S., Dekel, T.: Tokenflow: Consistent diffusion features for consistent video editing. CoRR  \textbf{abs/2307.10373} (2023)

\bibitem{DBLP:journals/corr/abs-2307-04725}
Guo, Y., Yang, C., Rao, A., Wang, Y., Qiao, Y., Lin, D., Dai, B.: Animatediff: Animate your personalized text-to-image diffusion models without specific tuning. CoRR  \textbf{abs/2307.04725} (2023)

\bibitem{DBLP:conf/iclr/HertzMTAPC23}
Hertz, A., Mokady, R., Tenenbaum, J., Aberman, K., Pritch, Y., Cohen{-}Or, D.: Prompt-to-prompt image editing with cross-attention control. In: The Eleventh International Conference on Learning Representations (2023)

\bibitem{ho2022imagen}
Ho, J., Chan, W., Saharia, C., Whang, J., Gao, R., Gritsenko, A., Kingma, D.P., Poole, B., Norouzi, M., Fleet, D.J., et~al.: Imagen video: High definition video generation with diffusion models. arXiv preprint arXiv:2210.02303  (2022)

\bibitem{DBLP:journals/jmlr/HoSCFNS22}
Ho, J., Saharia, C., Chan, W., Fleet, D.J., Norouzi, M., Salimans, T.: Cascaded diffusion models for high fidelity image generation. J. Mach. Learn. Res.  \textbf{23},  47:1--47:33 (2022)

\bibitem{DBLP:journals/corr/abs-2207-12598}
Ho, J., Salimans, T.: Classifier-free diffusion guidance. CoRR  \textbf{abs/2207.12598} (2022)

\bibitem{ho2022video}
Ho, J., Salimans, T., Gritsenko, A., Chan, W., Norouzi, M., Fleet, D.J.: Video diffusion models. In: NeurIPS (2022)

\bibitem{DBLP:conf/iclr/HuSWALWWC22}
Hu, E.J., Shen, Y., Wallis, P., Allen{-}Zhu, Z., Li, Y., Wang, S., Wang, L., Chen, W.: Lora: Low-rank adaptation of large language models. In: The Tenth International Conference on Learning Representations (2022)

\bibitem{DBLP:journals/tog/KastenOWD21}
Kasten, Y., Ofri, D., Wang, O., Dekel, T.: Layered neural atlases for consistent video editing. {ACM} Trans. Graph.  \textbf{40}(6),  210:1--210:12 (2021)

\bibitem{DBLP:conf/cvpr/KawarZLTCDMI23}
Kawar, B., Zada, S., Lang, O., Tov, O., Chang, H., Dekel, T., Mosseri, I., Irani, M.: Imagic: Text-based real image editing with diffusion models. In: {IEEE/CVF} Conference on Computer Vision and Pattern Recognition. pp. 6007--6017 (2023)

\bibitem{khachatryan2023text2video}
Khachatryan, L., Movsisyan, A., Tadevosyan, V., Henschel, R., Wang, Z., Navasardyan, S., Shi, H.: Text2video-zero: Text-to-image diffusion models are zero-shot video generators. arXiv preprint arXiv:2303.13439  (2023)

\bibitem{DBLP:journals/corr/KingmaW13}
Kingma, D.P., Welling, M.: Auto-encoding variational bayes. In: Bengio, Y., LeCun, Y. (eds.) 2nd International Conference on Learning Representations, {ICLR} 2014 (2014)

\bibitem{DBLP:journals/corr/abs-2305-01569}
Kirstain, Y., Polyak, A., Singer, U., Matiana, S., Penna, J., Levy, O.: Pick-a-pic: An open dataset of user preferences for text-to-image generation. arxiv  \textbf{abs/2305.01569} (2023)

\bibitem{DBLP:journals/corr/abs-2303-04761}
Liu, S., Zhang, Y., Li, W., Lin, Z., Jia, J.: Video-p2p: Video editing with cross-attention control. CoRR  \textbf{abs/2303.04761} (2023)

\bibitem{DBLP:conf/iclr/MengHSSWZE22}
Meng, C., He, Y., Song, Y., Song, J., Wu, J., Zhu, J., Ermon, S.: Sdedit: Guided image synthesis and editing with stochastic differential equations. In: The Tenth International Conference on Learning Representations, {ICLR} 2022 (2022)

\bibitem{DBLP:conf/cvpr/MokadyHAPC23}
Mokady, R., Hertz, A., Aberman, K., Pritch, Y., Cohen{-}Or, D.: Null-text inversion for editing real images using guided diffusion models. In: {IEEE/CVF} Conference on Computer Vision and Pattern Recognition. pp. 6038--6047 (2023)

\bibitem{DBLP:journals/corr/abs-2302-01329}
Molad, E., Horwitz, E., Valevski, D., Rav{-}Acha, A., Matias, Y., Pritch, Y., Leviathan, Y., Hoshen, Y.: Dreamix: Video diffusion models are general video editors. CoRR  \textbf{abs/2302.01329} (2023)

\bibitem{nichol2021glide}
Nichol, A., Dhariwal, P., Ramesh, A., Shyam, P., Mishkin, P., McGrew, B., Sutskever, I., Chen, M.: Glide: Towards photorealistic image generation and editing with text-guided diffusion models. arXiv preprint arXiv:2112.10741  (2021)

\bibitem{DBLP:conf/icml/NicholDRSMMSC22}
Nichol, A.Q., Dhariwal, P., Ramesh, A., Shyam, P., Mishkin, P., McGrew, B., Sutskever, I., Chen, M.: {GLIDE:} towards photorealistic image generation and editing with text-guided diffusion models. In: Chaudhuri, K., Jegelka, S., Song, L., Szepesv{\'{a}}ri, C., Niu, G., Sabato, S. (eds.) International Conference on Machine Learning, {ICML} 2022. Proceedings of Machine Learning Research, vol.~162, pp. 16784--16804

\bibitem{DBLP:journals/corr/abs-2303-08774}
OpenAI: {GPT-4} technical report. arXiv  \textbf{abs/2303.08774} (2023)

\bibitem{DBLP:journals/corr/abs-2308-07926}
Ouyang, H., Wang, Q., Xiao, Y., Bai, Q., Zhang, J., Zheng, K., Zhou, X., Chen, Q., Shen, Y.: Codef: Content deformation fields for temporally consistent video processing. CoRR  \textbf{abs/2308.07926} (2023). \doi{10.48550/ARXIV.2308.07926}

\bibitem{DBLP:conf/iccv/ParkSBBGSM21}
Park, K., Sinha, U., Barron, J.T., Bouaziz, S., Goldman, D.B., Seitz, S.M., Martin{-}Brualla, R.: Nerfies: Deformable neural radiance fields. In: 2021 {IEEE/CVF} International Conference on Computer Vision, {ICCV} 2021, Montreal, QC, Canada, October 10-17, 2021. pp. 5845--5854 (2021)

\bibitem{peebles2023scalable}
Peebles, W., Xie, S.: Scalable diffusion models with transformers. In: Proceedings of the IEEE/CVF International Conference on Computer Vision. pp. 4195--4205 (2023)

\bibitem{DBLP:journals/corr/abs-2307-01952}
Podell, D., English, Z., Lacey, K., Blattmann, A., Dockhorn, T., M{\"{u}}ller, J., Penna, J., Rombach, R.: {SDXL:} improving latent diffusion models for high-resolution image synthesis. CoRR  \textbf{abs/2307.01952} (2023)

\bibitem{podell2023sdxl}
Podell, D., English, Z., Lacey, K., Blattmann, A., Dockhorn, T., M{\"u}ller, J., Penna, J., Rombach, R.: Sdxl: improving latent diffusion models for high-resolution image synthesis. arXiv preprint arXiv:2307.01952  (2023)

\bibitem{DBLP:conf/iccv/QiCZLWSC23}
Qi, C., Cun, X., Zhang, Y., Lei, C., Wang, X., Shan, Y., Chen, Q.: Fatezero: Fusing attentions for zero-shot text-based video editing. In: {IEEE/CVF} International Conference on Computer Vision, {ICCV} 2023. pp. 15886--15896 (2023)

\bibitem{Gal2022image}
R, G., Y, A., Y, A.: An image is worth one word: Personalizing text-to-image generation using textual. arXiv preprint arXiv:2208.01618  (2022)

\bibitem{DBLP:conf/cvpr/RombachBLEO22}
Rombach, R., Blattmann, A., Lorenz, D., Esser, P., Ommer, B.: High-resolution image synthesis with latent diffusion models. In: {IEEE/CVF} Conference on Computer Vision and Pattern Recognition. pp. 10674--10685 (2022)

\bibitem{rombach2022high}
Rombach, R., Blattmann, A., Lorenz, D., Esser, P., Ommer, B.: High-resolution image synthesis with latent diffusion models. In: Proceedings of the IEEE/CVF conference on computer vision and pattern recognition. pp. 10684--10695 (2022)

\bibitem{ronneberger2015u}
Ronneberger, O., Fischer, P., Brox, T.: U-net: Convolutional networks for biomedical image segmentation. In: Medical Image Computing and Computer-Assisted Intervention--MICCAI 2015: 18th International Conference, Munich, Germany, October 5-9, 2015, Proceedings, Part III 18. pp. 234--241. Springer (2015)

\bibitem{DBLP:conf/cvpr/RuizLJPRA23}
Ruiz, N., Li, Y., Jampani, V., Pritch, Y., Rubinstein, M., Aberman, K.: Dreambooth: Fine tuning text-to-image diffusion models for subject-driven generation. In: {IEEE/CVF} Conference on Computer Vision and Pattern Recognition, {CVPR} 2023, Vancouver, BC, Canada, June 17-24, 2023. pp. 22500--22510 (2023)

\bibitem{saharia2022photorealistic}
Saharia, C., Chan, W., Saxena, S., Li, L., Whang, J., Denton, E.L., Ghasemipour, K., Gontijo~Lopes, R., Karagol~Ayan, B., Salimans, T., et~al.: Photorealistic text-to-image diffusion models with deep language understanding. Advances in Neural Information Processing Systems  \textbf{35},  36479--36494 (2022)

\bibitem{singer2022make}
Singer, U., Polyak, A., Hayes, T., Yin, X., An, J., Zhang, S., Hu, Q., Yang, H., Ashual, O., Gafni, O., et~al.: Make-a-video: Text-to-video generation without text-video data. arXiv preprint arXiv:2209.14792  (2022)

\bibitem{DBLP:conf/iclr/SongME21}
Song, J., Meng, C., Ermon, S.: Denoising diffusion implicit models. In: 9th International Conference on Learning Representations (2021)

\bibitem{DBLP:conf/cvpr/TumanyanGBD23}
Tumanyan, N., Geyer, M., Bagon, S., Dekel, T.: Plug-and-play diffusion features for text-driven image-to-image translation. In: {IEEE/CVF} Conference on Computer Vision and Pattern Recognition. pp. 1921--1930 (2023)

\bibitem{DBLP:journals/corr/abs-2308-06571}
Wang, J., Yuan, H., Chen, D., Zhang, Y., Wang, X., Zhang, S.: Modelscope text-to-video technical report. arXiv  \textbf{abs/2308.06571} (2023)

\bibitem{DBLP:conf/nips/Wang0TLCK19}
Wang, T., Liu, M., Tao, A., Liu, G., Catanzaro, B., Kautz, J.: Few-shot video-to-video synthesis. In: Wallach, H.M., Larochelle, H., Beygelzimer, A., d'Alch{\'{e}}{-}Buc, F., Fox, E.B., Garnett, R. (eds.) Advances in Neural Information Processing Systems 32: Annual Conference on Neural Information Processing Systems 2019, NeurIPS 2019. pp. 5014--5025 (2019)

\bibitem{DBLP:journals/corr/abs-2303-17599}
Wang, W., Xie, K., Liu, Z., Chen, H., Cao, Y., Wang, X., Shen, C.: Zero-shot video editing using off-the-shelf image diffusion models. arxiv  \textbf{abs/2303.17599} (2023)

\bibitem{wang2023videocomposer}
Wang, X., Yuan, H., Zhang, S., Chen, D., Wang, J., Zhang, Y., Shen, Y., Zhao, D., Zhou, J.: Videocomposer: Compositional video synthesis with motion controllability. arXiv preprint arXiv:2306.02018  (2023)

\bibitem{DBLP:journals/corr/abs-2401-07781}
Wu, J.Z., Fang, G., Wu, H., Wang, X., Ge, Y., Cun, X., Zhang, D.J., Liu, J., Gu, Y., Zhao, R., Lin, W., Hsu, W., Shan, Y., Shou, M.Z.: Towards {A} better metric for text-to-video generation. arxiv  \textbf{abs/2401.07781} (2024)

\bibitem{wu2023tune}
Wu, J.Z., Ge, Y., Wang, X., Lei, S.W., Gu, Y., Shi, Y., Hsu, W., Shan, Y., Qie, X., Shou, M.Z.: Tune-a-video: One-shot tuning of image diffusion models for text-to-video generation. In: Proceedings of the IEEE/CVF International Conference on Computer Vision. pp. 7623--7633 (2023)

\bibitem{DBLP:conf/iccv/WuGWLGSHSQS23}
Wu, J.Z., Ge, Y., Wang, X., Lei, S.W., Gu, Y., Shi, Y., Hsu, W., Shan, Y., Qie, X., Shou, M.Z.: Tune-a-video: One-shot tuning of image diffusion models for text-to-video generation. In: {IEEE/CVF} International Conference on Computer Vision. pp. 7589--7599 (2023)

\bibitem{DBLP:journals/corr/abs-2308-09710}
Xing, Z., Dai, Q., Hu, H., Wu, Z., Jiang, Y.: Simda: Simple diffusion adapter for efficient video generation. CoRR  \textbf{abs/2308.09710} (2023)

\bibitem{DBLP:conf/siggrapha/YangZLL23}
Yang, S., Zhou, Y., Liu, Z., Loy, C.C.: Rerender {A} video: Zero-shot text-guided video-to-video translation. In: {SIGGRAPH} Asia 2023 Conference Papers, {SA} 2023, Sydney, NSW, Australia. pp. 95:1--95:11 (2023)

\bibitem{yin2023dragnuwa}
Yin, S., Wu, C., Liang, J., Shi, J., Li, H., Ming, G., Duan, N.: Dragnuwa: Fine-grained control in video generation by integrating text, image, and trajectory. arXiv preprint arXiv:2308.08089  (2023)

\bibitem{DBLP:journals/tmlr/YuWVYSW22}
Yu, J., Wang, Z., Vasudevan, V., Yeung, L., Seyedhosseini, M., Wu, Y.: Coca: Contrastive captioners are image-text foundation models. Trans. Mach. Learn. Res.  (2022)

\bibitem{VBLIP}
Yu, K.P.: Videoblip: Supercharged blip-2 that can handle videos. (2023), \url{https://github.com/yukw777/VideoBLIP}

\bibitem{DBLP:journals/corr/abs-2306-10012}
Zhang, K., Mo, L., Chen, W., Sun, H., Su, Y.: Magicbrush: {A} manually annotated dataset for instruction-guided image editing. arxiv  \textbf{abs/2306.10012} (2023)

\bibitem{DBLP:journals/corr/abs-2311-04145}
Zhang, S., Wang, J., Zhang, Y., Zhao, K., Yuan, H., Qin, Z., Wang, X., Zhao, D., Zhou, J.: I2vgen-xl: High-quality image-to-video synthesis via cascaded diffusion models. arxiv  \textbf{abs/2311.04145} (2023)

\bibitem{zhang2023controlvideo}
Zhang, Y., Wei, Y., Jiang, D., Zhang, X., Zuo, W., Tian, Q.: Controlvideo: Training-free controllable text-to-video generation. arXiv preprint arXiv:2305.13077  (2023)

\bibitem{DBLP:conf/eccv/ZhuoWLW022}
Zhuo, L., Wang, G., Li, S., Wu, W., Liu, Z.: Fast-vid2vid: Spatial-temporal compression for video-to-video synthesis. In: Avidan, S., Brostow, G.J., Ciss{\'{e}}, M., Farinella, G.M., Hassner, T. (eds.) Computer Vision - {ECCV} 2022 - 17th European Conference, Tel Aviv, Israel, October 23-27, 2022, Proceedings, Part {XV}. vol. 13675, pp. 289--305 (2022)

\end{thebibliography}
\end{document}